%% file: main.tex
\documentclass[10pt,twocolumn,letterpaper]{article}
\usepackage[pagenumbers]{cvpr} 
\usepackage{float}
\usepackage{xcolor}

\definecolor{cvprblue}{rgb}{0.21,0.49,0.74}
\usepackage[pagebackref=true,breaklinks=true,letterpaper=true,colorlinks,bookmarks=false]{hyperref}
\usepackage{multirow}
\usepackage{multicol}
\usepackage[T1]{fontenc}

\title{T-Rex: Counting by Visual Prompting}

\author{Qing Jiang$^{1,2*}$ , Feng Li$^{1,3*}$ , Tianhe Ren$^{1}$ , Shilong Liu$^{1,4*}$ , Zhaoyang Zeng$^{1}$ \\ Kent Yu$^{1}$ , Lei Zhang$^{1\dagger}$ \\
$^1$International Digital Economy Academy (IDEA) \\
$^2$SCUT \qquad$^3$HKUST \qquad$^4$Tsinghua \\
{\tt\small mountchicken@outlook.com , fliay@connect.ust.hk , lius120@mails.tsinghua.edu.cn} \\ {\tt\small \{rentianhe, zengzhaoyang, kentyu, leizhang\}@idea.edu.cn} \\
\url{trex-counting.github.io}
}
\vspace{-2mm}
\begin{document}

\twocolumn[{
\maketitle\centering
\captionsetup{type=figure}
\includegraphics[width=0.9\textwidth]{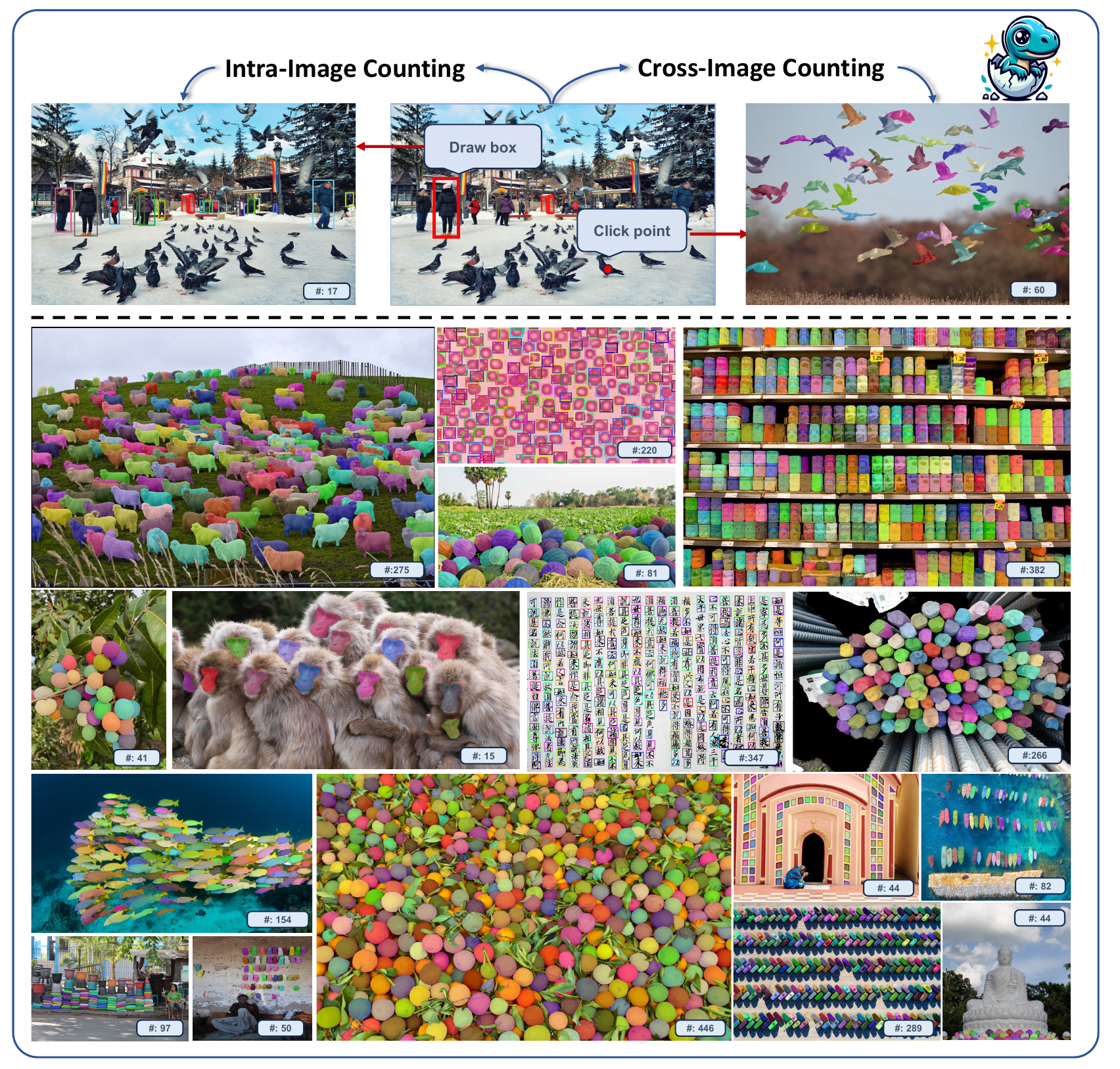}\vspace{-2mm}
\captionof{figure}{We introduce an interactive object counting model, T-Rex. Given boxes or points specified on the reference image, T-Rex can detect all instances on the target image that exhibit similar pattern with the specified object, which are then summed to obtain the counting result. We use SAM\cite{Kirillov_2023_ICCV} to generate mask prompted on the detected boxes by T-Rex for better visualization.}
\label{fig:teaser}
}]

\setcounter{footnote}{0}

\renewcommand{\thefootnote}{\fnsymbol{footnote}}
\footnotetext[1]{\emph{This work was done when Qing Jiang, Feng Li, and Shilong Liu were interns at IDEA.}}

\footnotetext[2]{\emph{Project lead.}}

\renewcommand{\thefootnote}{\arabic{footnote}}
\setcounter{footnote}{0}

\maketitle

\input{sec/01_abstract}
\input{sec/02_introduction}
\input{sec/03_related_work}
\input{sec/04_method}
\input{sec/05_experiments}

\input{sec/06_potential_applications}

\end{document}

%% file: sec/01_abstract.tex
\begin{abstract}
We introduce T-Rex\footnote{\emph{We named our model "T-Rex" to signify its visual detection capability, akin to the Tyrannosaurus Rex's exceptional visual acuity, which is estimated to be 13 times better than that of a human. \href{https://en.wikipedia.org/wiki/Tyrannosaurus}{Wikipedia link.}}}\footnote{\emph{Technical Report. Work in progress.}}, an interactive object counting model designed to first detect and then count any objects. We formulate object counting as an open-set object detection task with the integration of visual prompts. Users can specify the objects of interest by marking points or boxes on a reference image, and T-Rex then detects all objects with a similar pattern. Guided by the visual feedback from T-Rex, users can also interactively refine the counting results by prompting on missing or falsely-detected objects. T-Rex has achieved state-of-the-art performance on several class-agnostic counting benchmarks. To further exploit its potential, we established a new counting benchmark encompassing diverse scenarios and challenges. Both quantitative and qualitative results show that T-Rex possesses exceptional zero-shot counting capabilities. We also present various practical application scenarios for T-Rex, illustrating its potential in the realm of visual prompting.
\end{abstract}

%% file: sec/02_introduction.tex
\section{Introduction}\label{sec:introduction}

Computer Vision is currently undergoing a revolution dominated by foundation models \cite{Kirillov_2023_ICCV, jia2021scaling, liu2022swin, wang2023internimage, radford2021learning, cheng2022masked, ramesh2022hierarchical, rombach2022high} and multi-modal large-language models \cite{chen2023pali, lv2023kosmos, wang2023visionllm, wang2022git, yu2022coca, alayrac2022flamingo, liu2023visual, zhang2023meta, girdhar2023imagebind, openai, peng2023kosmos}. These models have demonstrated remarkable performance across a wide spectrum of tasks, including segmentation \cite{cheng2022masked, Kirillov_2023_ICCV, li2023mask}, object detection \cite{zhang2022dino, liu2023grounding, zong2023detrs}, understanding \cite{li2023blip, peng2023kosmos, openai, liu2023visual, alayrac2022flamingo, wang2023visionllm} , and generation \cite {ramesh2022hierarchical, rombach2022high, zhang2023adding}. However, among these tasks, the critical task of object counting has received relatively less attention.

Object counting, the task of estimating the number of specific objects present in an image, is in high demand across numerous practical fields, such as transportation, agriculture, industry, biology, etc. Existing solutions for object counting can be broadly categorized into four types:

\begin{itemize}
  \item \textbf{As density map regression task.} A common approach \cite{lempitsky2010learning, djukic2023low, huang2023interactive, nguyen2022few, ranjan2021learning, jiang2023clip} is to regress a 2D density map, the summation of which is used as the counting result. Although effective, the \textit{less intuitive visualization} of the density map \cite{huang2023interactive} makes it difficult  for users to assess the accuracy of the counting results.
  \item \textbf{As closed-set detection task.} Another straightforward solution involves employing a closed-set detector (E.g. YOLO \cite{redmon2016you}) to detect objects, where the summation of the number of detected boxes serves as the counting result. However, \textit{limited by the fixed categories}, this method requires data re-collection and re-training efforts for novel categories, which is time-consuming and labor-intensive.
  \item \textbf{As open-vocabulary detection task.} To overcome the limitations of closed-set detection methods, an alternative approach is to adapt open vocabulary detector (E.g. Grounding DINO \cite{liu2023grounding}) to detect arbitrary objects through text prompts. However, the task of counting poses a significant challenge as many objects \textit{do not have concise textual descriptions}, making object specification by text difficult.
  \item \textbf{As MLLM QA task.} Multi-modal Large Language Models (MLLM) can also be used for object counting through question answering \cite{yang2023dawn, li2023otterhd}. However, the issue of hallucination \cite{ji2023survey} in multi-modal LLMs affects  the \textit{confidence level} of their counting results, as users may be skeptical of a numerical output from an MLLM without additional supporting evidence.
\end{itemize}

\begin{figure}[t]
\centering
\includegraphics[width=0.95\linewidth]{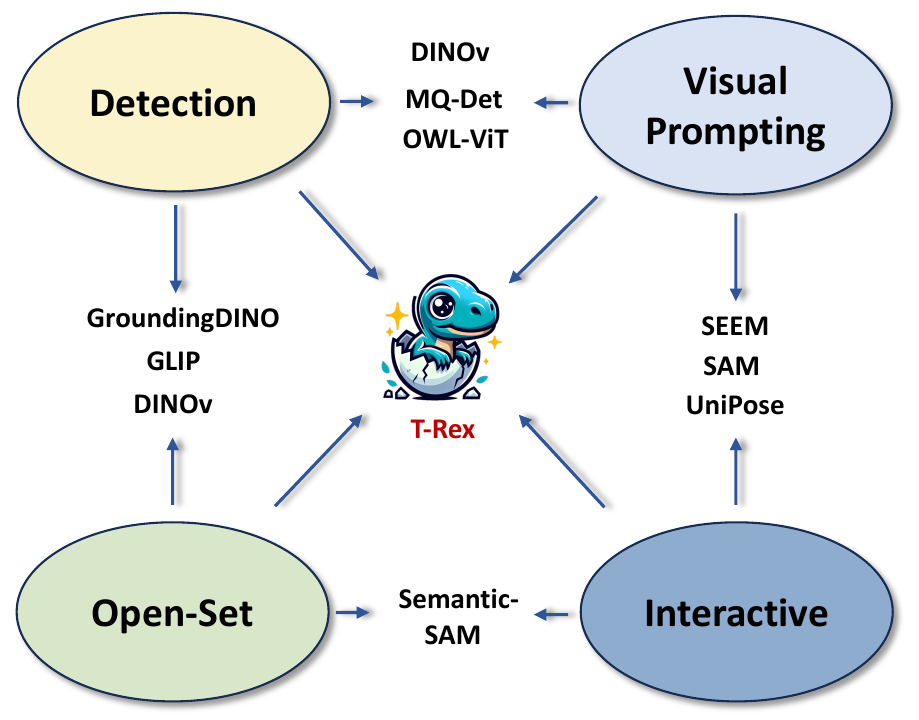}\vspace{-1mm}
\caption{T-Rex is an object counting model, which is characterized by four features: detection-based, visual promptable, interative, and open-set. Listed methods are: Grounding DINO \cite{liu2023grounding}, GLIP \cite{li2022grounded},Semantic-SAM \cite{li2023semantic}, SEEM \cite{zou2023segment}, SAM \cite{Kirillov_2023_ICCV},   UniPose \cite{yang2023unipose}, MQ-Det \cite{xu2023multi}, OWL-ViT \cite{minderer2022simple}, DINOv \cite{li2023visual}.}
\label{fig:scope} 
\vspace{-1mm}
\end{figure}

By highlighting  the limitations of existing counting solutions, we argue that a practical counting system should possess the following four properties: a) \textbf{Intuitive Visual Feedback}: It should provide highly interpretable visual feedback (e.g. bounding box), allowing users to verify the accuracy of the counting results. b) \textbf{Open-Set}:  It should be capable of counting any objects, without constraints on predefined categories. c) \textbf{Visual Promptable}: It should allow users to specify the objects for counting through visual examples, given the limitation of text to discripe various objects. d) \textbf{Interactive}: It should enable users to actively participate in the counting process to correct errors made by the model.

Guided by this design philosophy, we develop an detection-based counting model, called T-Rex, as shown in Fig. \ref{fig:scope}. Users can specify the object of interest by marking boxes or points on the reference image. T-Rex, in return, detects all instances with a similar pattern in the target image, and the cumulative sum of the detected boxes represents the counting result. With the visual feedback from T-Rex, users can interactively add additional prompts on missed or falsely-detected objects. This interactive process allows for continual refinement over T-Rex's prediction, empowering users to confidently access the accuracy of the counting results. Notably, this interactive process remains fast and resource-efficient, as each round of interaction only requires forwarding the decoder of T-Rex.

T-Rex has achieved state-of-the-art results on two counting benchmarks \cite{ranjan2021learning, nguyen2022few}. To further measure its potential, we introduce a new counting benchmark, CA-44, which comprises 44 datasets across eight domains, presenting diverse and formidable challenges. Our experimental findings demonstrate that T-Rex possesses strong zero-shot counting capabilities and can achieve good performance in various scenarios. Finally, we explore a wide range of application scenes of T-Rex.  With its versatile counting capabilities and interactive features, T-Rex has the potential to make substantial contributions to various domains, such as retail, transportation, agriculture, industry, livestock, medicine, etc.

%% file: sec/03_related_work.tex
\section{Related Works}
\label{sec:related_works}

\textbf{Object Counting. }Methods for object counting can be divided into class-specific and class-agnostic. Class-specific approaches typically use object detection models to count specific categories, such as people \cite{abousamra2021localization, zhang2015cross}, cars \cite{hsieh2017drone, mundhenk2016large}, or animals \cite{arteta2016counting}. These methods are limited to predefined classes and require additional data labeling for new categories, which is time-consuming and labor-intensive. To overcome the limitations of close-set object detector, class-agnostic methods \cite{djukic2023low, huang2023interactive, nguyen2022few, ranjan2021learning, jiang2023clip} have been developed to regress density map \cite{lempitsky2010learning} based on the correlation feature between image and a few visual exemplars. However, these techniques often lack an intuitive visualization, making it difficult for users to verify the model's accuracy.

\noindent\textbf{Interactive Models. }Interactive models have shown significant promise in aligning human intentions within the field of computer vision. SAM\cite{Kirillov_2023_ICCV} presents an interactive segmentation model capable of accommodating point, box,
and text-based input, achieving remarkable zero-shot segmentation by leveraging an large-scale dataset SA-1B\cite{Kirillov_2023_ICCV}. SEEM \cite{zou2023segment} and Semantic-SAM \cite{li2023semantic} have extended to more
general segmentation models that can output the semantic
meaning for the segmented mask. In contrast, the field of
interactive object detection is less explored. One concurrent work DINOv~\cite{li2023visual}, explores visual in-context prompting in both referring and general segmentation. However, its performance is sub-optimal in processing images with densely packed objects, and it lacks the capability for multi-round interactions necessary for correcting erroneous detection results.

%% file: sec/04_method.tex
\begin{figure*}[t]\centering
\includegraphics[width=\linewidth]{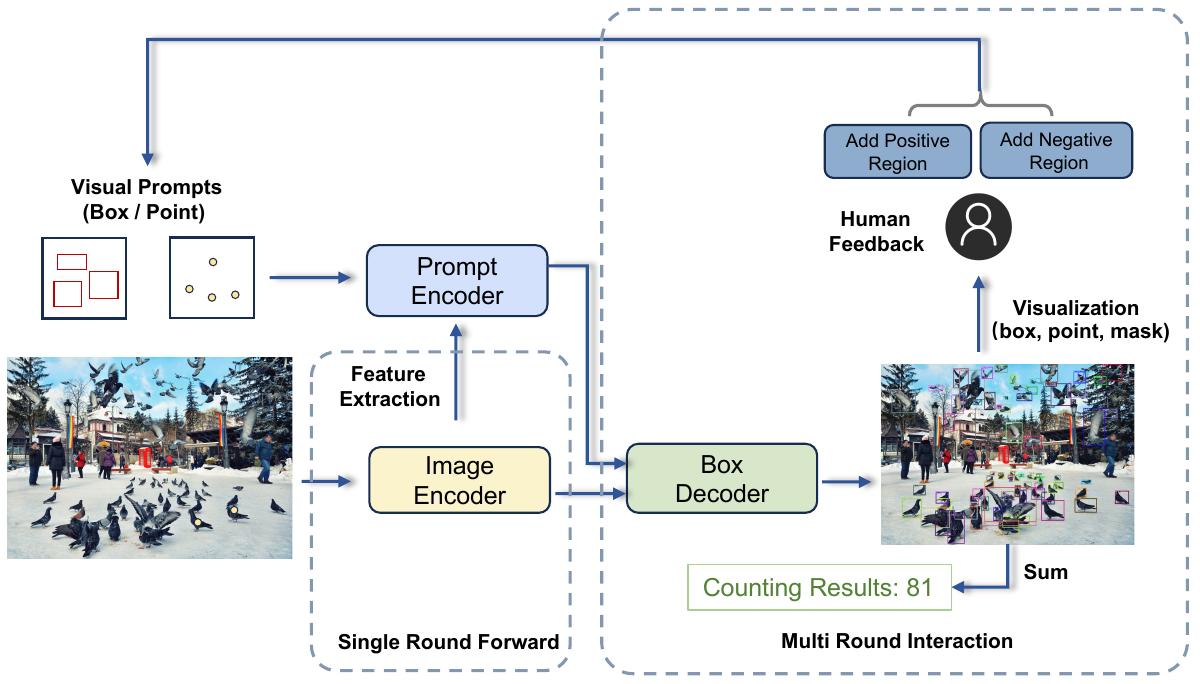}\vspace{-1mm}
\caption{Overview of the T-Rex model. T-Rex is a detection-based model comprising an image encoder to extract image feature, a prompt encoder to encode visual prompts (points or boxes) provided by users, and a box decoder to output the detected boxes.}
\label{fig:model_diagram}
\vspace{-1mm}
\end{figure*}

\section{Overview of T-Rex}
We briefly introduce the T-Rex model. T-Rex comprises of three components, including an image encoder, a prompt encoder and a box decoder, as illustrated in Fig. \ref{fig:model_diagram}. Given a target image input $\mathbf{I}_{tgt}$ and optionally a reference image input $\mathbf{I}_{ref}$ (the target image can also serve as the reference image in the absence of a separate reference image), the image encoder first extracts the visual features $\mathbf{E}_{tgt}$, $\mathbf{E}_{ref}$. Then, using the user-drawn boxes or points as prompts $\mathbf{P}$ for the target object on the reference image, the prompt encoder extracts the encoded visual prompt $\mathbf{P}_{enc}$ from the reference image feature $\mathbf{E}_{ref}$. Finally, the box decoder combines the target image feature $\mathbf{E}_{tgt}$ and the encoded visual prompt $\mathbf{P}_{enc}$ as inputs, outputting detected boxes $\mathbf{B}$ along with their associated confidence scores $\mathbf{S}$. A predetermined score threshold is applied to filter the detected boxes, and the remaining boxes are summed to produce the final object count.
\begin{equation}
    \mathbf{E}_{tgt}=\mathbf{ImageEncoder}\left(\mathbf{I}_{tgt}\right)
\end{equation}
\begin{equation}
    \mathbf{E}_{ref }=\mathbf{ImageEncoder}\left(\mathbf{I}_{ref}\right) 
\end{equation}
\begin{equation}
    \mathbf{P}_{enc}=\mathbf{PromptEncoder}\left(\mathbf{P}, \mathbf{E}_{ref}\right) 
\end{equation}
\begin{equation}
    \mathbf{B, S}=\mathbf{BoxDecoder}\left(\mathbf{P}_{enc}, \mathbf{E}_{tgt}\right) 
\end{equation}
\begin{equation}
    \mathbf{\#Count}=\sum{\mathbf{ThreshFilter}(\mathbf{B}, \mathbf{S})}
\end{equation}

\subsection{Workflows}
\begin{figure*}[t]\centering
\includegraphics[width=\linewidth]{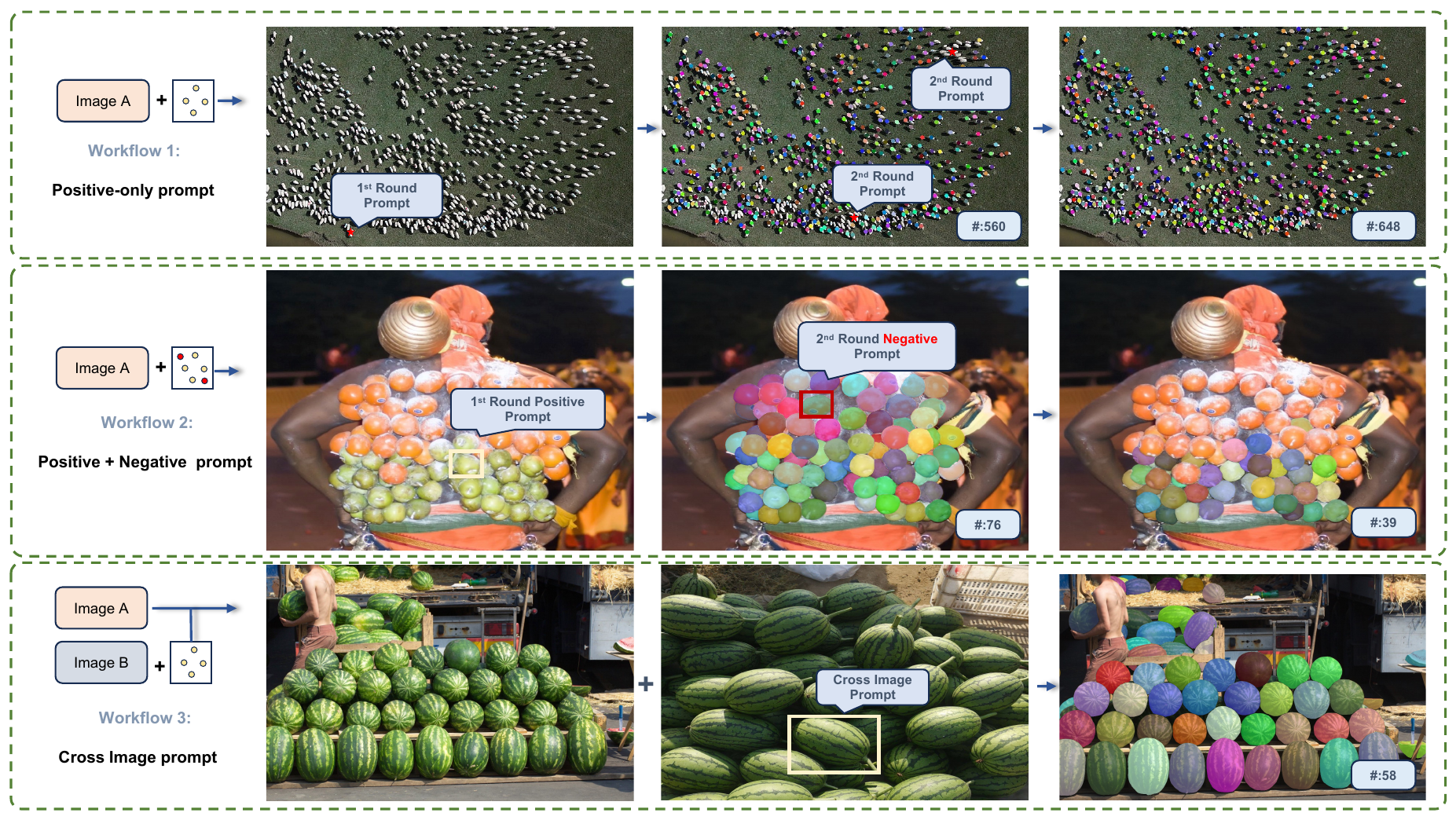}\vspace{-1mm}
\caption{T-Rex offers three major interactive workflows, which are applicable to most scenarios in real-world applications.}
\label{fig:cat_workflow}
\vspace{-1mm}
\end{figure*}

\begin{figure*}[t]\centering
\includegraphics[width=0.99\linewidth]{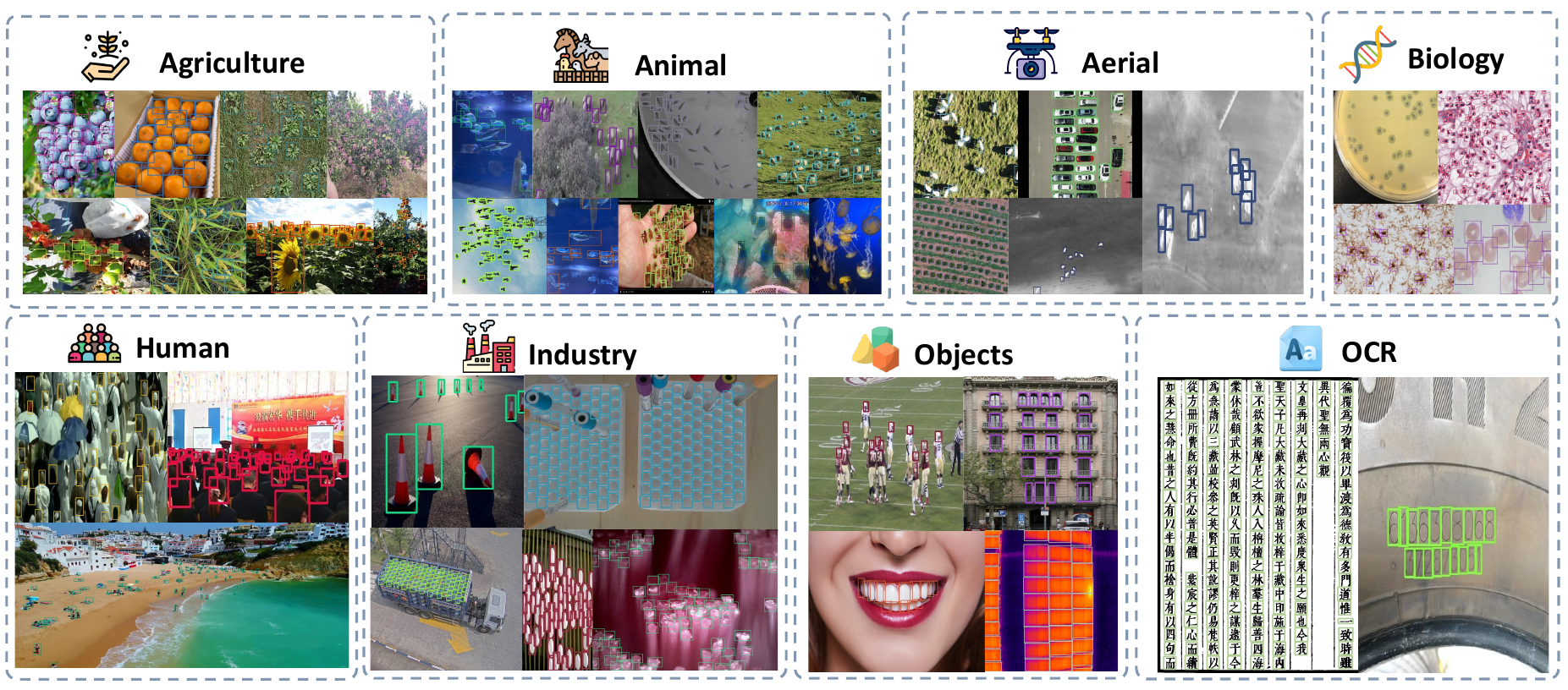}\vspace{-1mm}
\caption{An Overview of the proposed CA-44 benchmark. CA-44 consists of 44 datasets across eight domains and mainly comprises images with small and densely packed objects.}
\label{fig:ca_benchmark}
\vspace{-1mm}
\end{figure*}

T-Rex offers three major interactive workflows as shown in Fig. \ref{fig:cat_workflow}. We explain each workflow and its application below.

\textbf{Positive-only Prompt Mode. }In most counting scenarios, users typically only 
 need to click once or draw one box, and T-Rex can effectively detect all objects with a similar pattern. However, in cases involving dense and small objects, a single round of prompting may be insufficient. In such cases, users have the option to incorporate additional prompts to the missed region, based on the visual feedback from T-Rex. This iterative refinement approach allows for more accurate counting results.

\textbf{Positive with Negative Prompt Mode. }In scenarios where interference from other similar objects is present, T-Rex may generate falsely detected boxes. As illustrated in Fig. \ref{fig:cat_workflow}, when a prompt is directed at a green apple, T-Rex might erroneously detect the orange tangerine due to the strong geometric resemblance between these two object types. In such cases, users can rectify the counting result by adding negative prompts to the falsely detected object. 

\textbf{Cross-Image Prompt Mode. }T-Rex additionally offers support for cross-image counting, allowing for the combination of different reference and target images. This functionality proves especially useful in automatic annotation scenarios, where users need to prompt only once on one image, and T-Rex can automatically annotate other images that exhibit a similar object pattern to the prompted image.

\subsection{Discussion}
T-Rex is essentially an open-set object detection model. Compared with open-vocabulary object detectors \cite{liu2023grounding, li2022grounded, minderer2022simple, zhou2022detecting} that rely on text-based prompt, T-Rex adopts visual prompt instead. Since in many real-world counting applications, text descriptions may not sufficiently capture all object details, and employing visual prompts offers a more direct and versatile alternative.

In the context of the object counting task, a paramount consideration revolves around the need for highly reliable prediction from the model. Given that the counting results are represented as statistical values, even a minor discrepancy in the predicted value signifies a failure counting. Hence, we design T-Rex to be interactive, allowing users to iteratively rectify counting results based on the visual feedback, thus enhancing the counting accuracy. Regarding the structure of model, T-Rex requires only a single forward pass through the Image Encoder, while subsequent rounds of interaction involve only the Prompt Encoder and Box Decoder. This streamlined approach ensures that the entire interaction process remains lightweight and fast.

\section{Count Anything Benchmark}

To conduct a holistic performance evaluation of the T-Rex model, we develop a new object counting benchamrk named CA-44. This benchmark includes a total of 44 datasets covering eight distinct domains, as depicted in Fig. \ref{fig:ca_benchmark}.

\textbf{Dataset Distribution. }The majority of the datasets included in CA-44 were collected from \href{https://universe.roboflow.com/}{Roboflow} and underwent additional filtering procedures applied. For instance, we eliminated images containing fewer than 10 instances, since object counting is mostly focused on dense scenes. The composition of the CA-44 dataset is detailed in Table \ref{tab:dataset_dis_ca}.

\begin{table}[]
  \centering
  \resizebox{0.8\columnwidth}{!}{%
    \begin{tabular}{crrr}
        \hline
        Category    & \# Datasets & \# Images & \multicolumn{1}{l}{\# Instances} \\ \hline
        Industrial  & 6           & 1,781     & 114,688                         \\
        Object      & 4           & 3,644     & 94,328                          \\
        Biology     & 4           & 1,834     & 63,235                           \\
        OCR         & 2           & 305       & 86,986                           \\
        Animal      & 8           & 4,674     & 122,069                          \\
        Human       & 3           & 1,119     & 44,212                           \\
        Aerial      & 5           & 5,011     & 156,824                          \\
        Agriculture & 12          & 11,717    & 488,719                          \\ \hline
        Total       & 44          & 30,085    & 1,171,061                         \\ \hline
        \end{tabular}
  }
  \caption{Dataset distribution of CA-44.}
  \label{tab:dataset_dis_ca}
\end{table}

\begin{figure}[t]
\includegraphics[width=\linewidth]{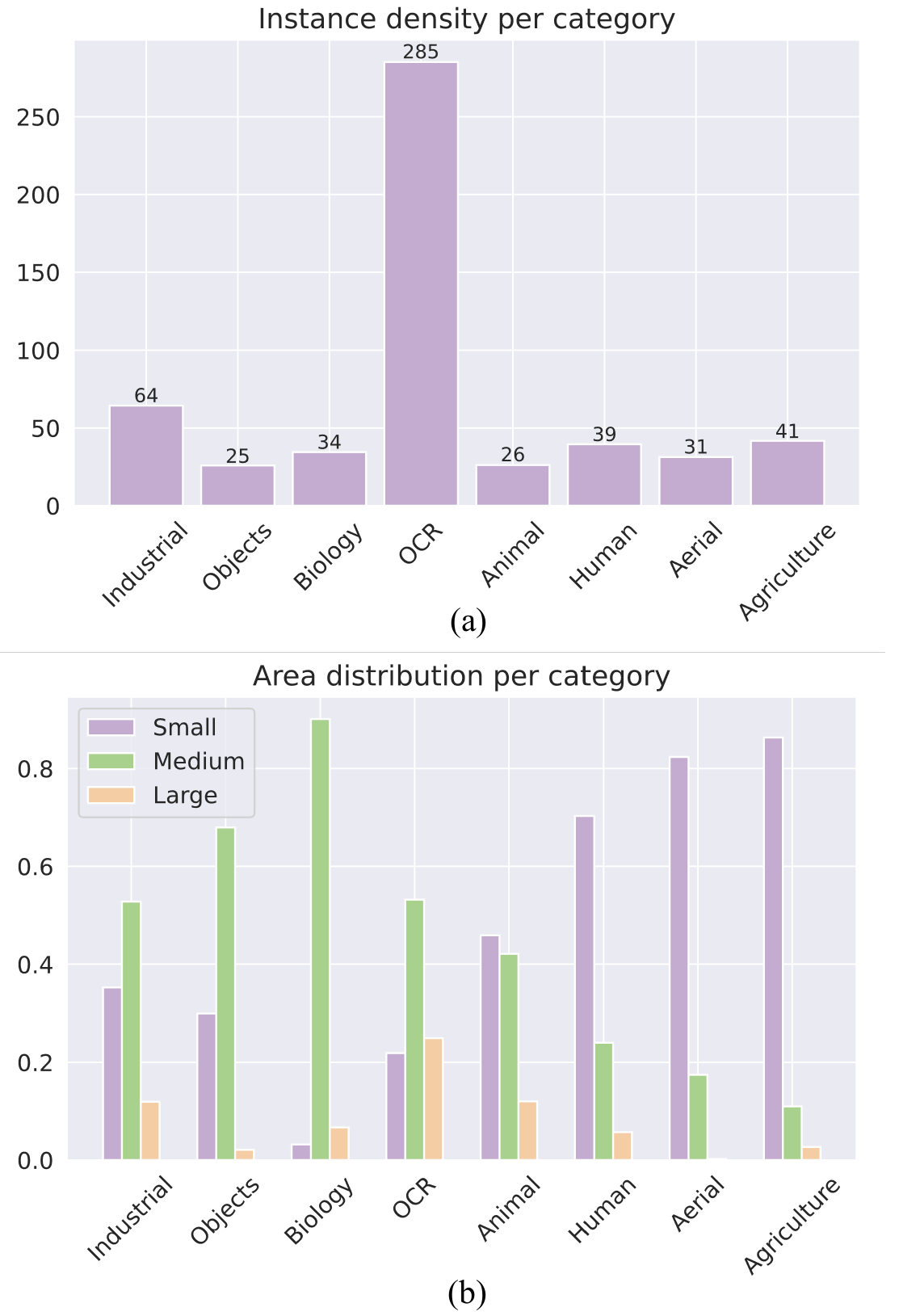}\vspace{-1mm}
\caption{Statistics of the CA-44 benchmark, highlighting the prevalence of small and dense objects. Object size categorization follows the COCO dataset \cite{lin2014microsoft} , where objects with an area smaller than $32^2$ are classified as small, those with an area between $32^2$ and $96^2$ as medium, and objects with area greater than $96^2$ as large.}
\label{fig:ca_statistic} 
\vspace{-1mm}
\end{figure}

\textbf{Statistics. }As illustrated in Fig. \ref{fig:ca_statistic}, the CA-44 benchmark primarily features images with small and densely packed objects. These characteristics reflect the common attributes of scenes in the object counting domain.

%% file: sec/05_experiments.tex

\section{Experiment Results}
\textbf{Settings. }In addition to our proposed CA-44 benchmark, we also conduct evaluations on the commonly-used counting dataset FSC147 \cite{ranjan2021learning} and the more challenging dataset FSCD-LVIS \cite{nguyen2022few}. FSC147 comprises 147 categories of objects and 1190 images in the test set and FSCD-LVIS comprises 377 categories and 1014 images in the test set. Both two datasets provide three bounding boxes of exemplar objects for each image, which we will use as the visual prompt for T-Rex.

\textbf{Metrics. }We adopt two metrics for evaluation: the Mean Average Error (MAE) metric, a widely employed standards in object counting and the Normalized Mean Average Error (NMAE) metric for more intuitive results. The mathematical expressions for the two metrics are as follows:
\begin{equation}
    \mathrm{MAE}=\frac{1}{J} \sum_{j=1}^J\left|c_j^*-c_j\right|
\end{equation}
\vspace{-1mm}
\begin{equation}
    \mathrm{NMAE}=\frac{1}{J} \sum_{j=1}^J \frac{\left|c_j^*-c_j\right|}{c_j^*}
\end{equation}
where $J$ represents the total number of test images, $c^*_j$ and $c_j$ denote the ground truth (GT) and the predicted number of objects for image $j$, respectively.

\subsection{Results on FSC147 and FSCD-LVIS}
The results on FSC147 and FSCD-LVIS are presented in Table \ref{tab:fsc147_oneshot}, Table \ref{tab:fsc147_threeshot}, and Table \ref{tab:fscd_lvis_threeshot}. T-Rex demonstrates state-of-the-art performance when compared to other density map regression-based methods, in both one-shot and three-shot settings. Beyond competitive performance, T-Rex also provides a user-friendly interactive counting interface. As a detection-based method, T-Rex offers intuitive visual feedback, allowing users to iteratively refine counting results and make informed judgments regarding their completeness. This interactive process enables T-Rex to achieve high reliability, contrasting with the less robust density map regression methods.

\begin{table}[]
  \centering
  \resizebox{0.7\columnwidth}{!}{%
    \begin{tabular}{clcccc}
        \hline
        Type                          & Method                                      & MAE$\downarrow$ \\ \hline
        \multirow{5}{*}{Density Map}  & FamNet \cite{ranjan2021learning}            & 26.76                   \\
                                      & BMNet+ \cite{shi2022represent}              & 16.89                    \\
                                      & LaoNet \cite{lin2021object}                 & 15.78                    \\
                                      & CountTR \cite{liu2022countr}                & 12.06                   \\
                                      & LOCA \cite{djukic2023low}                   & 12.53        \\ \hline
        \multicolumn{1}{c}{Detection} & T-Rex (Ours)                                  &\textbf{10.59}              \\ \hline
        \end{tabular}
  }
  \caption{One-shot counting evaluation on FSC147 test-set. One-shot indicates that each image utilizes one examplar box as visual prompt.}
  \label{tab:fsc147_oneshot}
\end{table}

\begin{table}[]
  \centering
  \resizebox{0.75\columnwidth}{!}{%
    \begin{tabular}{clcccc}
        \hline
        Type                          & Method                                      & MAE$\downarrow$ \\ \hline
        \multirow{5}{*}{Density Map}  & FamNet  \cite{ranjan2021learning}          & 22.08     \\
                                      & BMNet+ \cite{shi2022represent}               & 14.62         \\
                                      & SAFECount \cite{you2023few}                  & 14.32        \\
                                      & CountTR  \cite{liu2022countr}               & 11.95        \\
                                      & LOCA \cite{djukic2023low}                   & 10.79        \\ \hline
        \multicolumn{1}{c}{Detection} & T-Rex (Ours)                                  & \textbf{8.72}& \\ \hline
        \end{tabular}
  }
  \caption{Three-shot counting evaluation on FSC147 test-set. Three-shot indicates that each image utilizes three examplar box as visual prompts.}
  \label{tab:fsc147_threeshot}
\end{table}

\begin{table}[]
  \centering
  \resizebox{\columnwidth}{!}{%
    \begin{tabular}{clcccc}
        \hline
        Type                          & Method                                      & MAE$\downarrow$          & NMAE$\downarrow$  \\ \hline
        \multirow{2}{*}{Density Map}  & FamNet \cite{ranjan2021learning}            & 22.08                    &  1.82                  \\
                                      & Counting-DETR \cite{nguyen2022few}          & 18.51                    &  0.45                  \\ \hline
        \multicolumn{1}{c}{Detection} & T-Rex (Ours)                                  & \textbf{12.87}           &  \textbf{0.27}         \\ \hline
        \end{tabular}
  }
  \caption{Three-shot counting evaluation on FSCD-LVIS test-set.}
  \label{tab:fscd_lvis_threeshot}
\end{table}

\begin{figure}[t]
  \centering
  \includegraphics[scale=0.42]{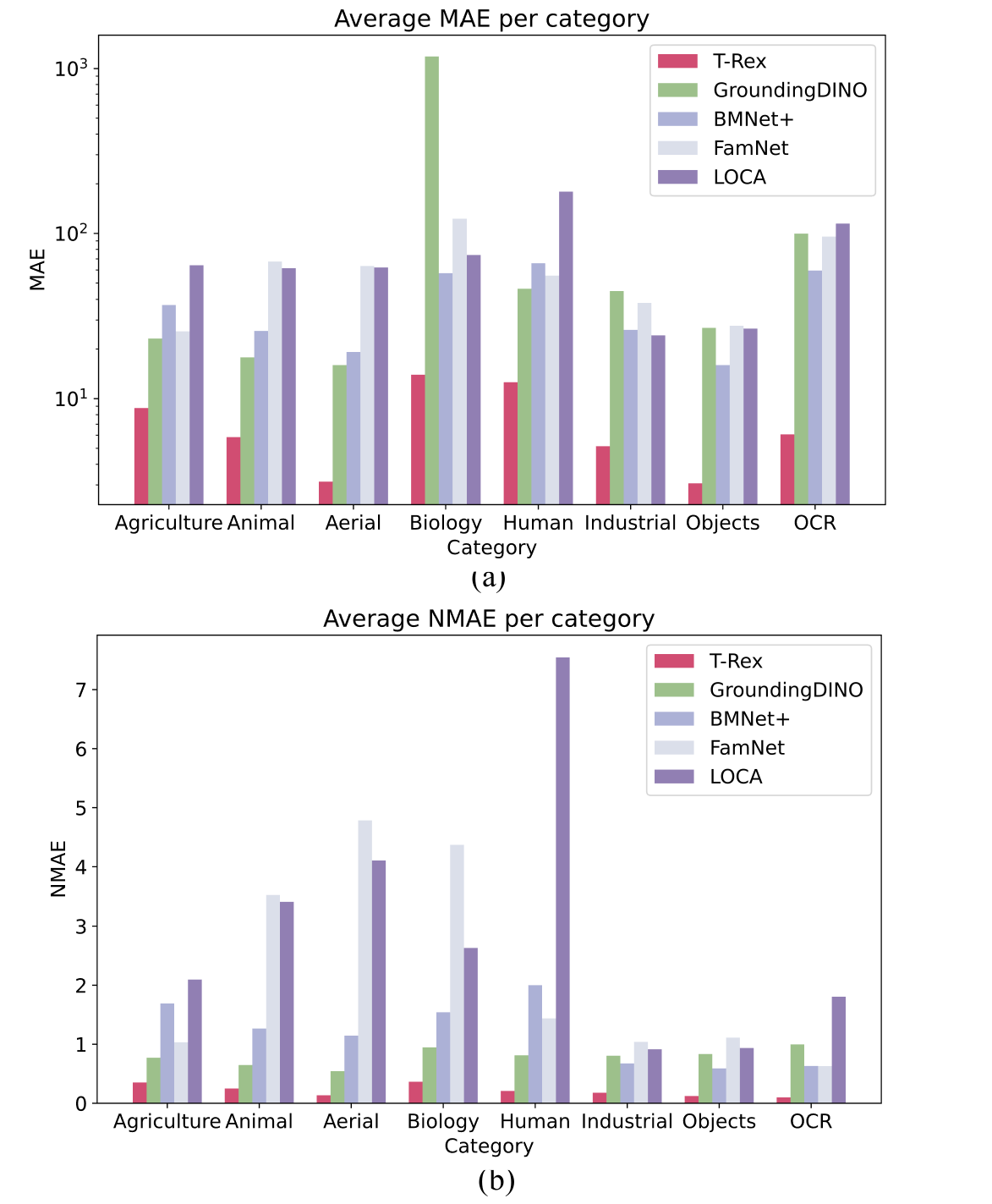}
  \caption{Results on Full CA-44. We compare T-Rex with open-vocabulary detector Grounding DINO \cite{liu2023grounding} and  density map regression-based methods \cite{shi2022represent, ranjan2021learning, djukic2023low}.}
  \label{fig:ca44_result}
  \vspace{-0.5em}
\end{figure}

\begin{figure}[t]
  \centering
  \includegraphics[scale=0.7]{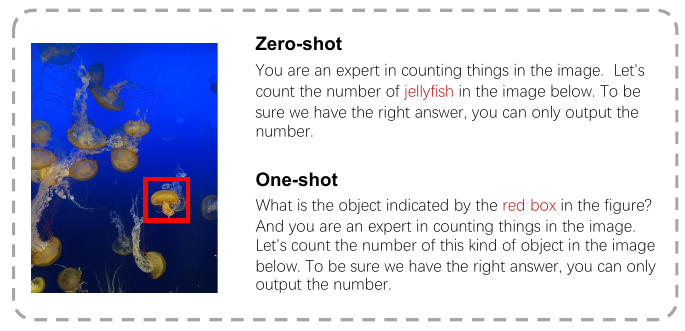}
  \caption{Prompts that are used to evaluate the counting ability of GPT-4V.}
  \label{fig:gpt4v_prompt}
  \vspace{-0.5em}
\end{figure}

\begin{figure}[t]
  \centering
  \includegraphics[scale=0.40]{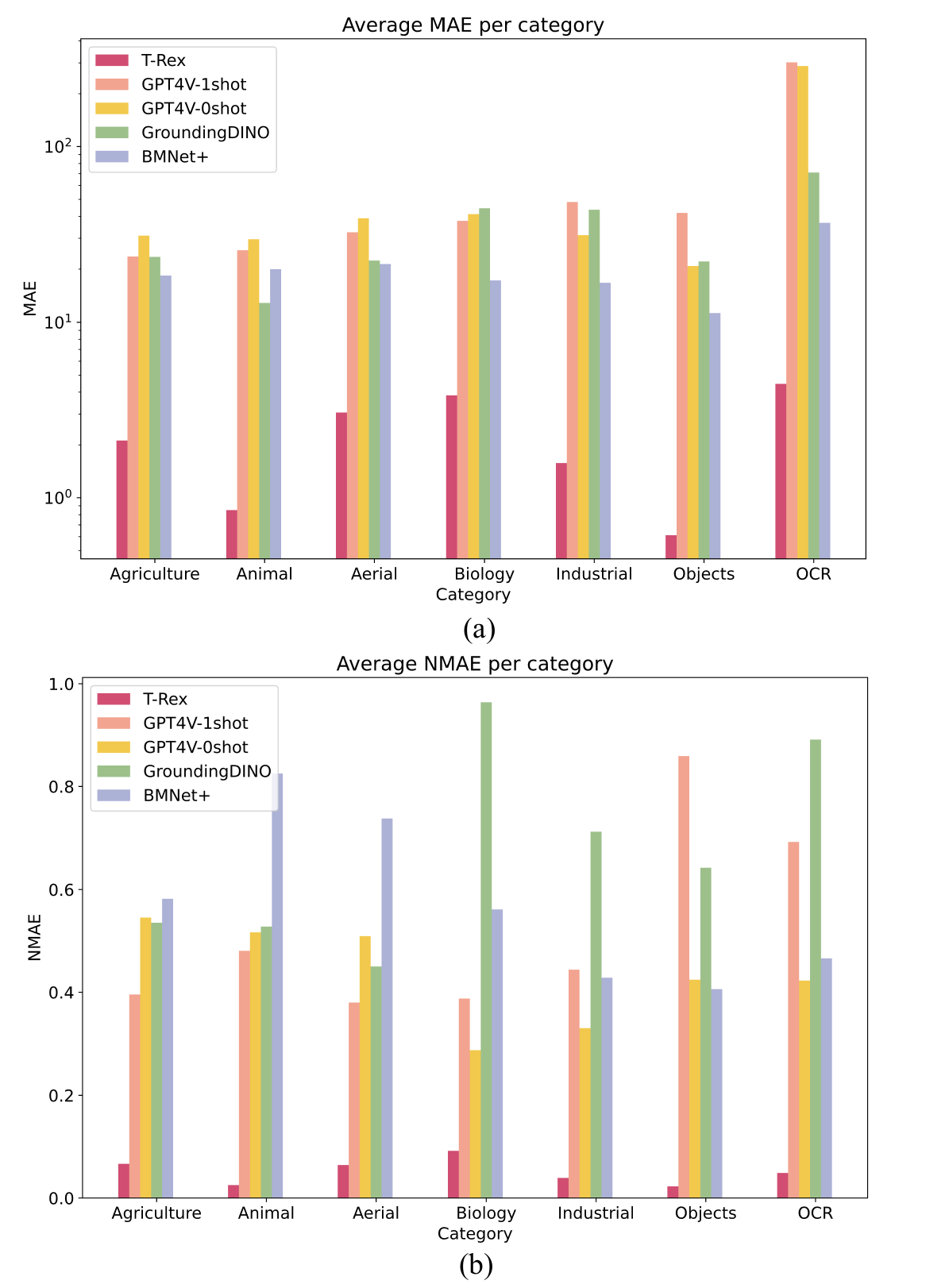}
  \caption{Results on subset of CA-44, sampling a total of 100 images from each category. GPT4V-0shot denotes zero-shot approach where we inform GPT-4V about the objects in the image and instruct it to count them. GPT4V-1shot denotes the one-shot approach where we annotate the objects in the image with bounding boxes and request GPT-4V to count them. }
  \label{fig:ca44_100_result}
  \vspace{-0.5em}
\end{figure}

\subsection{Results on CA-44}
Results on CA-44 are visually presented in Fig. \ref{fig:ca44_result}. T-Rex significantly outperforms other methods, including the open-vocabulary detector Grounding DINO \cite{liu2023grounding} and various density map regression-based methods, across all categories within CA-44. This showcases T-Rex's  exceptional zero-shot counting capabilities and its adaptness at tackling challenges spanning diverse domains. In comparison to the open-ocabulary detector Grounding DINO, T-Rex with visual prompt is more competitive. This underscores the limitations of text in providing sufficient descriptions, highlighting the significance of introducing visual prompts as a method. 

We have also conducted a comparison with the state-of-the-art multi-modality model, GPT-4V \cite{openai}, which has previously demonstrated counting capabilities \cite{yang2023dawn}. For an optimized evaluation, we selected a total of 100 images from CA-44 for testing. We tested GPT-4V in two settings: a zero-shot approach, where we inform GPT-4V about the objects in the image for counting, and a one-shot approach, where we annotate the objects in the image with bounding boxes for GPT-4V to count. The One-shot approach is similar to T-Rex in that both methods rely on visual prompts. Prompts used are detailed in Fig. \ref{fig:gpt4v_prompt}. Results presented in Fig. \ref{fig:ca44_100_result} illustrate that T-Rex outperforms GPT-4V in counting accuracy, suggesting that while large multi-modality models perform well in understanding tasks, they may still lack the nuanced capabilities for accurate object perception.

%% file: sec/06_potential_applications.tex

\section{Conclusion}
In this paper, we have introduced T-Rex, an innovative model for interactive object counting, characterized by its ability to detect and count objects using visual prompt. T-Rex represents a significant advancement in visual prompting methodologies within computer vision, paralleling the successes observed in NLP with LLMs facilitating Human-AI interactions through text prompts. This parallel suggests vast possibilities, that the application of visual prompts in computer vision could herald a comparable breakthrough like those in NLP.

\section{Limitations}
\begin{figure*}[t]\centering
\includegraphics[width=0.99\linewidth]{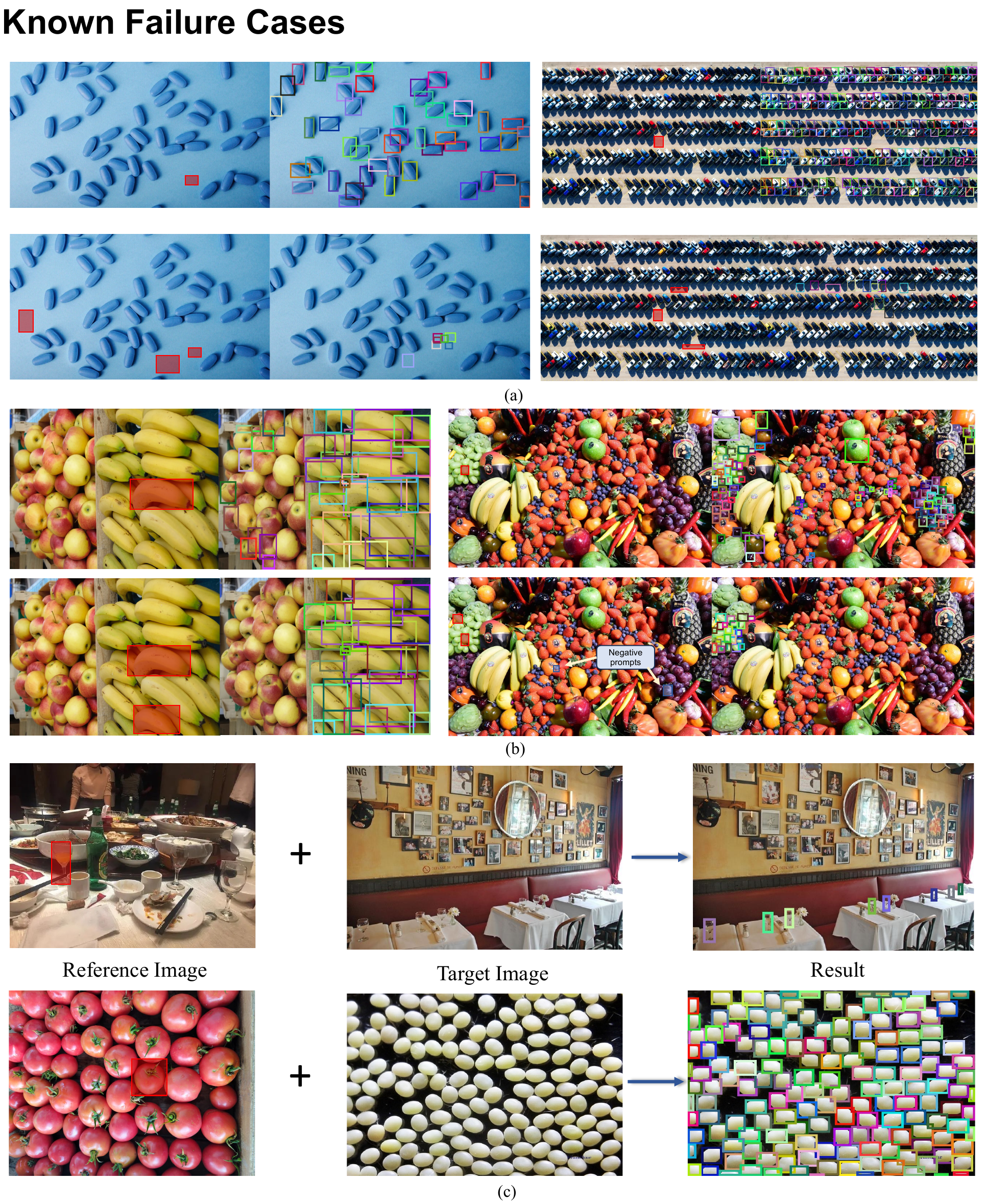}\vspace{-1mm}
\caption{Failure Cases in T-Rex: (a)In single-target scenes, if only a single prompt is introduced at the background, T-Rex will still detect dense objects in the figure. (b) In dense multi-object scenes, T-Rex can produce false detections, requiring multiple prompts or negative prompts for correction. (c) During cross-image workflows, T-Rex risks over-fitting in single-target dense scenes, leading to erroneous detections like mistaking silkworm eggs for tomatoes.}
\label{fig:faliure_case}
\vspace{-1mm}
\end{figure*}

T-Rex , in its intial version, has several limitations. We list a series of failure cases in Fig. \ref{fig:faliure_case}, where T-Rex may perform poorly. \textbf{Single-Target Scenes}. When only a single prompt is used against the background, T-Rex tends to misidentify dense objects clusters. \textbf{Dense Multi-Object Scenes}. T-Rex struggles in scenes with densely populated multi-object types, often leading to false detections. Addressing this issue may require either multiple iterations of prompting or the use of negative prompts. \textbf{Cross-Image Workflow}. A notable limitation emerges in cross-image workflow, especially when T-Rex is applied to scenes with a single target. In such scenarios, there is a significant risk of over-fitting, where T-Rex tends to ignore the user's prompt on the referece image. For example, even when prompted on tomatoes, T-Rex may still detect silkworm eggs. We will continue to improve the performance and robustness of T-Rex.

\section{Acknowledgments}
We would like to express our deepest gratitude to multiple teams within IDEA for their substantial support in the T-Rex project. We sincerely appreciate the CVR team, whose essential contributions and technical expertise were pivotal in realizing the project's goals. We thank Wei Liu, Xiaohui Wang, and Yakun Hu from the Product team for their strategic insights and innovative input in the development of the demo.  Appreciation is also extended to Yuanhao Zhu and Ce Feng from the Front-End team for their technical excellence and dedication. The robust solutions provided by Weiqiang Hu, Xiaoke Jiang, and Zhiqiang Li from the Back-End team were also crucial in supporting the project's infrastructure. We also thank Jie Yang for helpful discussion and Ling-Hao Chen for helping in video demos.

\section{Potential Applications}
In this section, we explore the application of T-Rex in various fields. As a detection-based model, it can be used as an object counter, as well as an automatic-annotation tool.

\begin{figure*}[t]\centering
\includegraphics[width=0.99\linewidth]{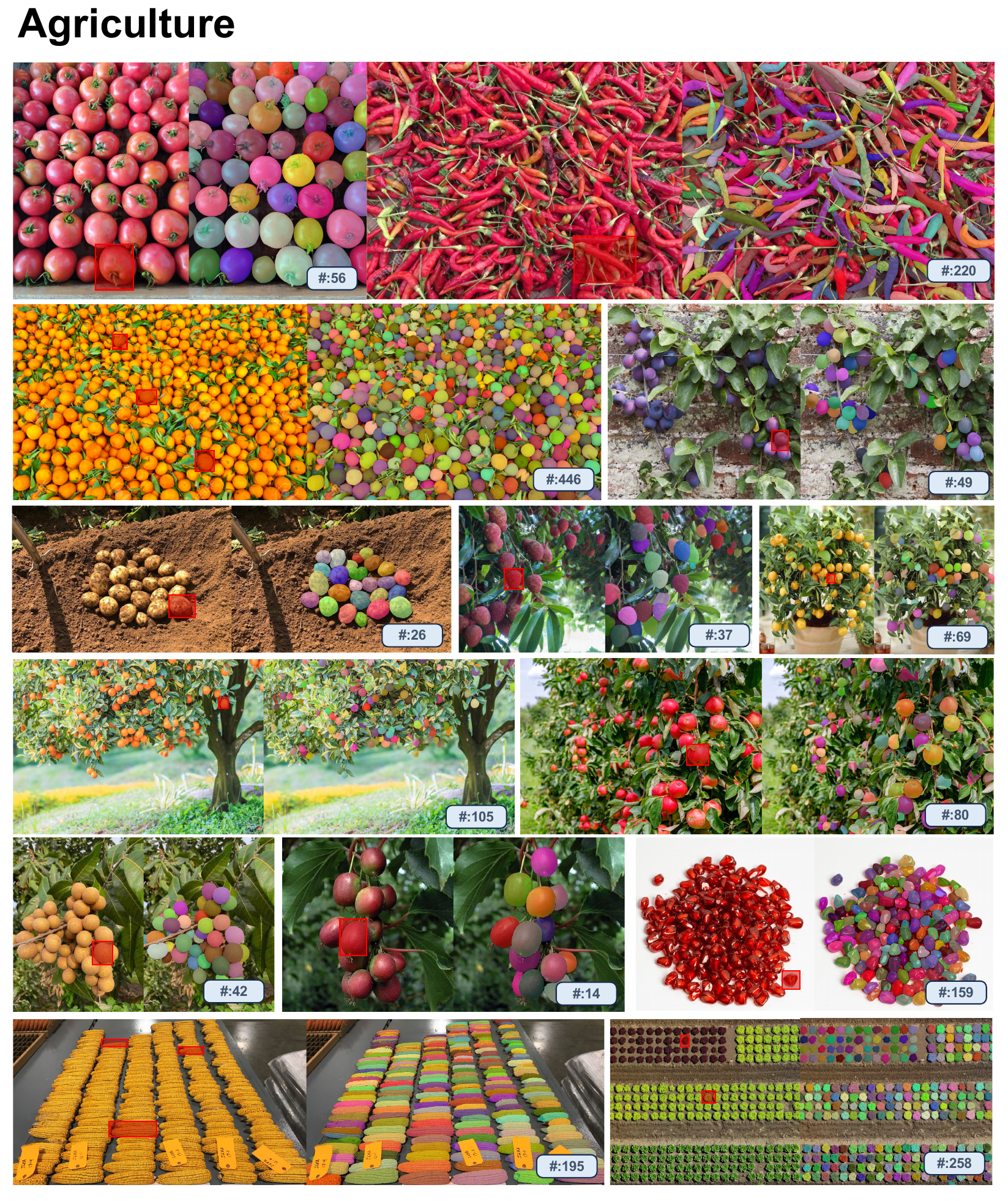}\vspace{-1mm}
\caption{T-Rex applied to agriculture. To more intuitively visualize overlapping and dense objects, we prompt SAM with the boxes predicted by T-Rex to obtain segmentation masks.}
\label{fig:agriculture}
\vspace{-1mm}
\end{figure*}

\begin{figure*}[t]\centering
\includegraphics[width=0.99\linewidth]{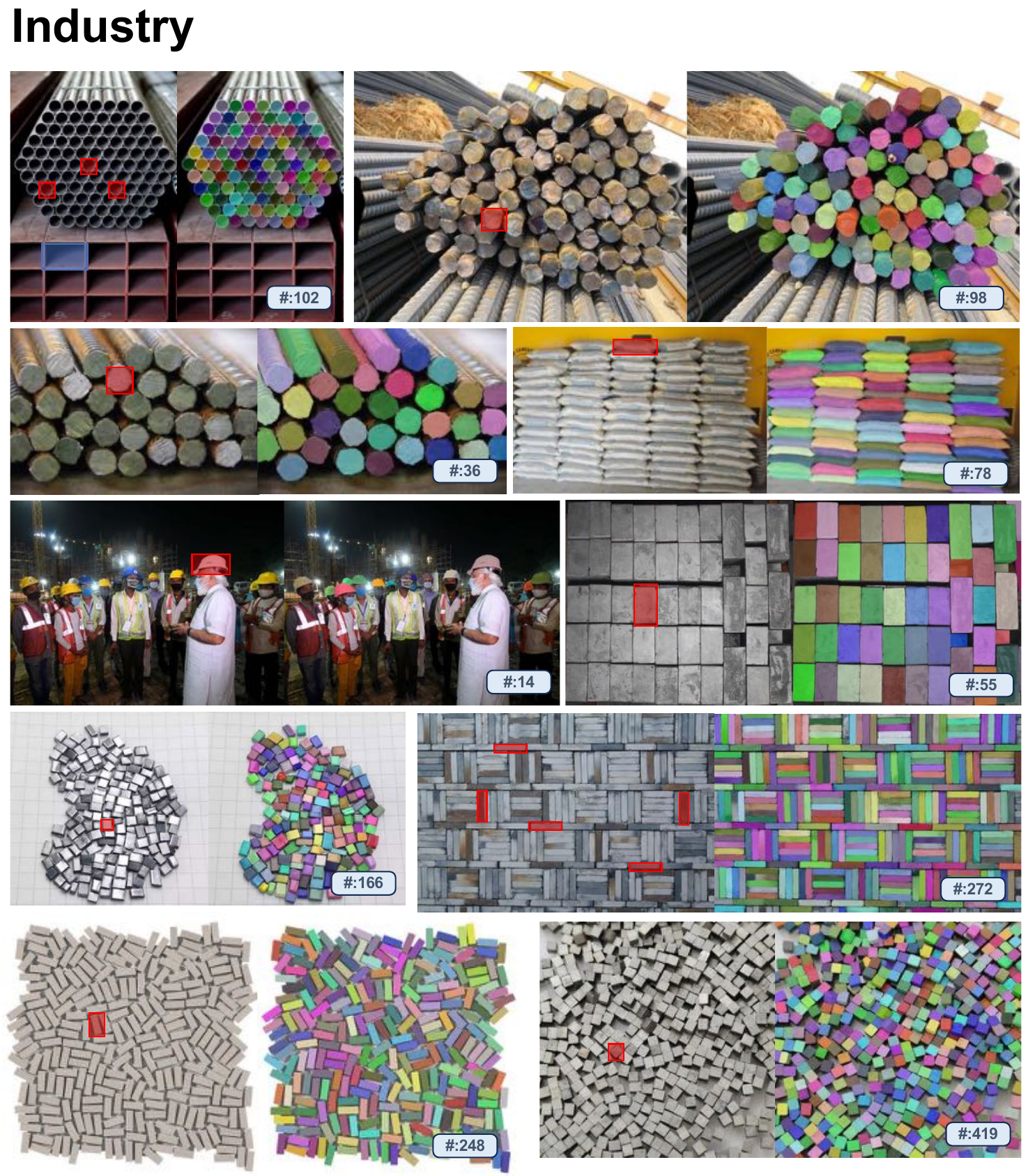}\vspace{-1mm}
\caption{T-Rex applied to industry. To more intuitively visualize overlapping and dense objects, we prompt SAM with the boxes predicted by T-Rex to obtain segmentation masks.}
\label{fig:industry1}
\vspace{-1mm}
\end{figure*}

\begin{figure*}[t]\centering
\includegraphics[width=0.99\linewidth]{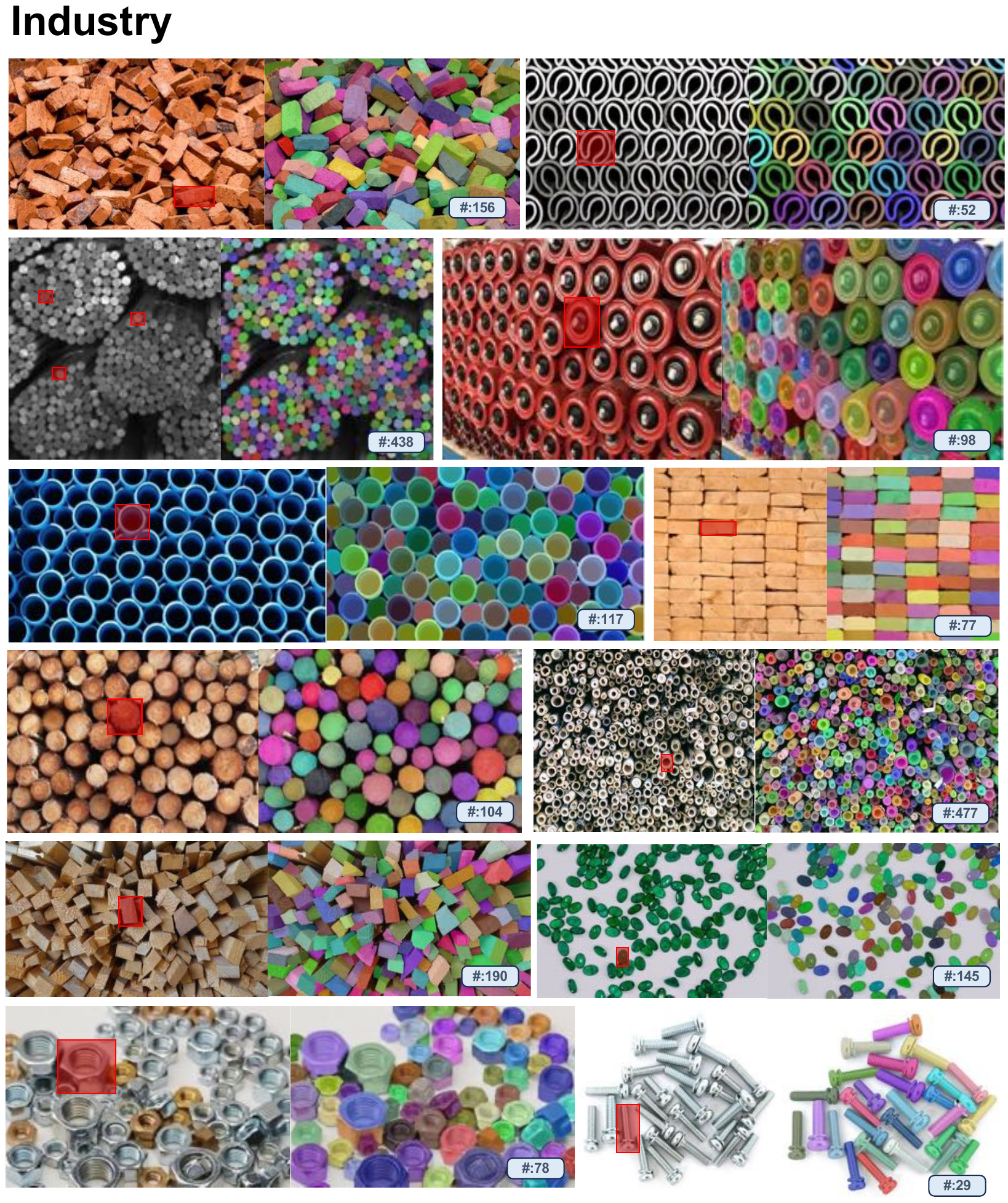}\vspace{-1mm}
\caption{T-Rex applied to industry. To more intuitively visualize overlapping and dense objects, we prompt SAM with the boxes predicted by T-Rex to obtain segmentation masks.}
\label{fig:industry2}
\vspace{-1mm}
\end{figure*}

\begin{figure*}[t]\centering
\includegraphics[width=0.99\linewidth]{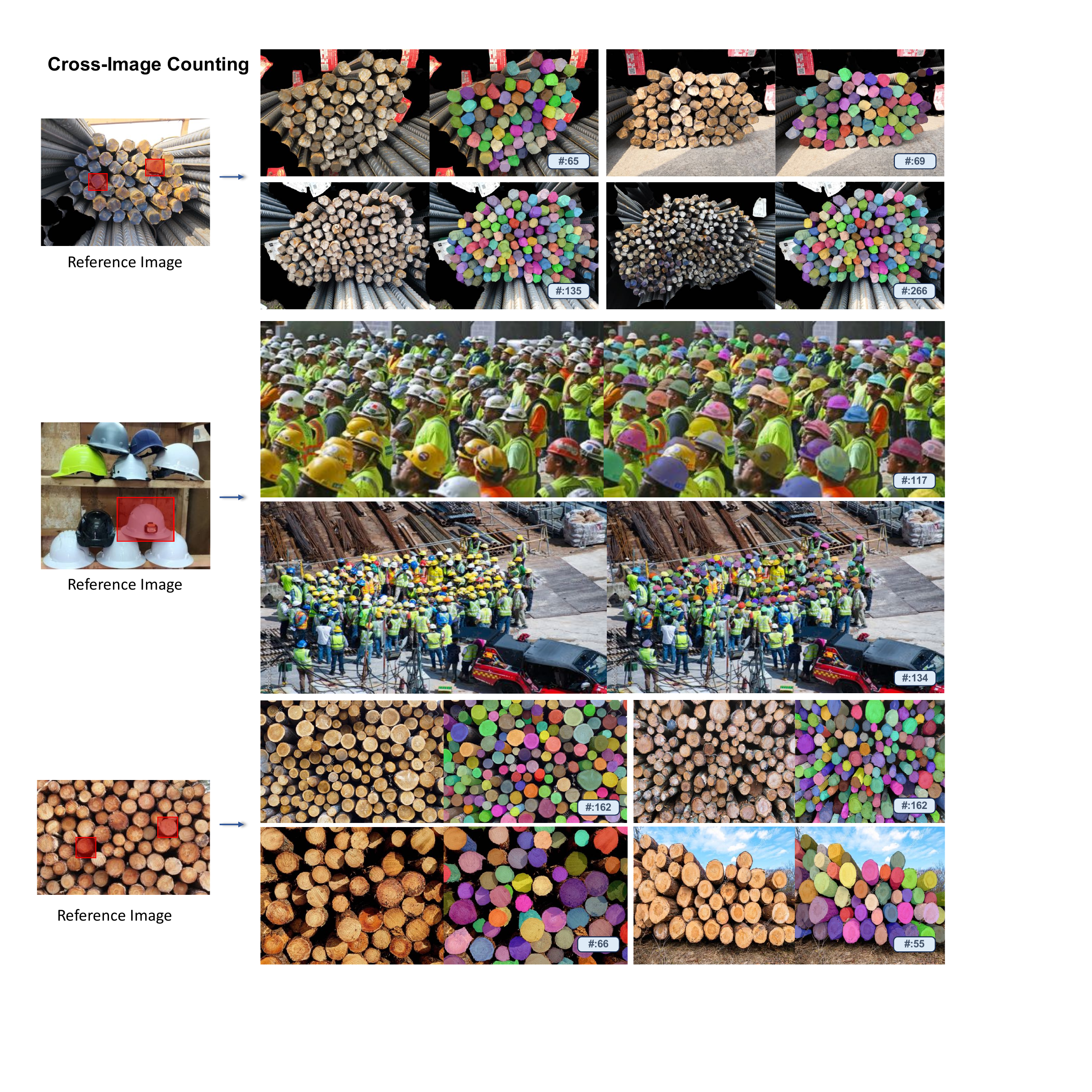}\vspace{-1mm}
\caption{T-Rex applied to industry in cross-image reference mode. To more intuitively visualize overlapping and dense objects, we prompt SAM with the boxes predicted by T-Rex to obtain segmentation masks.}
\label{fig:industry3}
\vspace{-1mm}
\end{figure*}

\begin{figure*}[t]\centering
\includegraphics[width=0.99\linewidth]{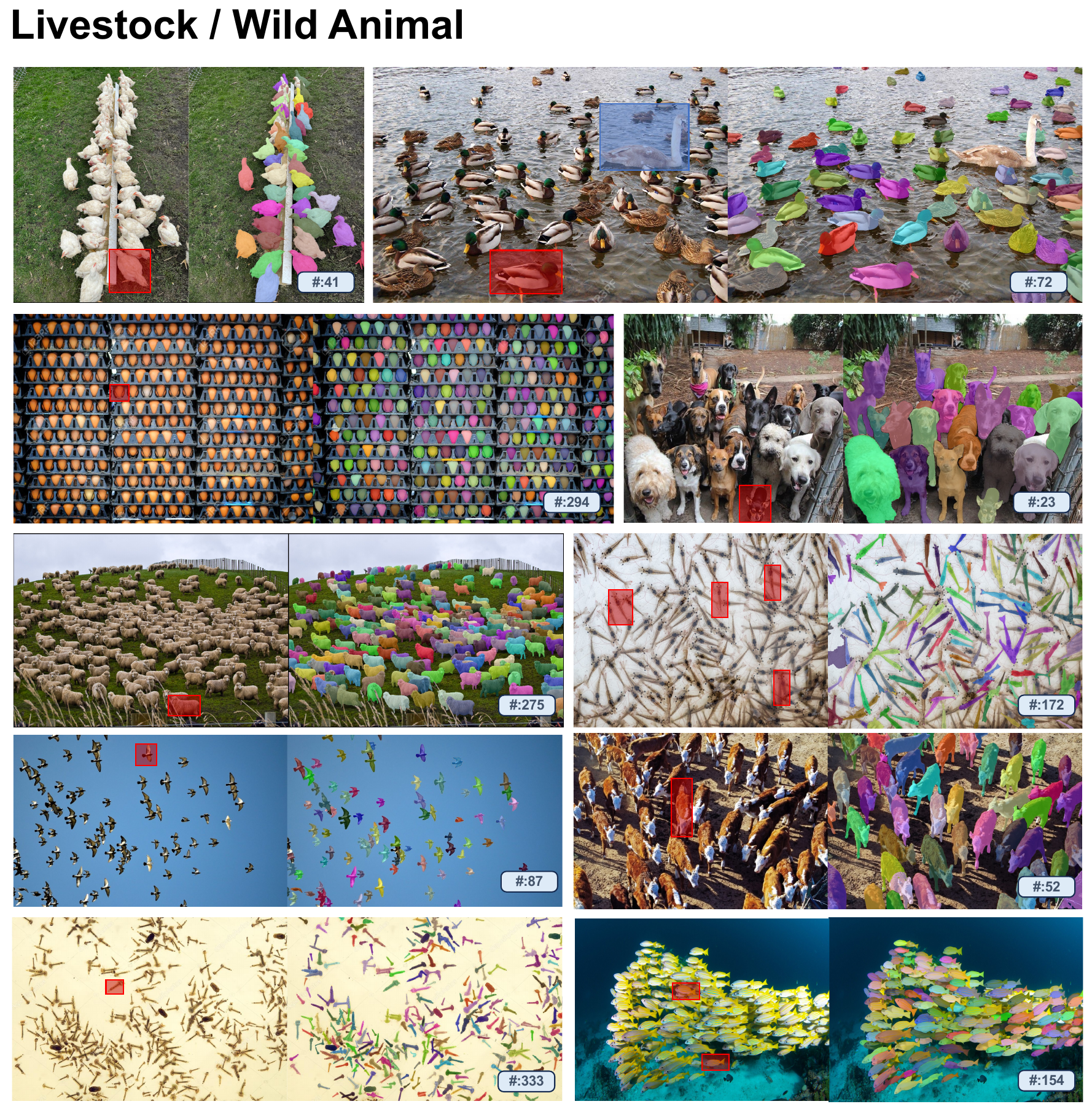}\vspace{-1mm}
\caption{T-Rex applied to livestock and wild animal. To more intuitively visualize overlapping and dense objects, we prompt SAM with the boxes predicted by T-Rex to obtain segmentation masks.}
\label{fig:animal}
\vspace{-1mm}
\end{figure*}

\begin{figure*}[t]\centering
\includegraphics[width=0.99\linewidth]{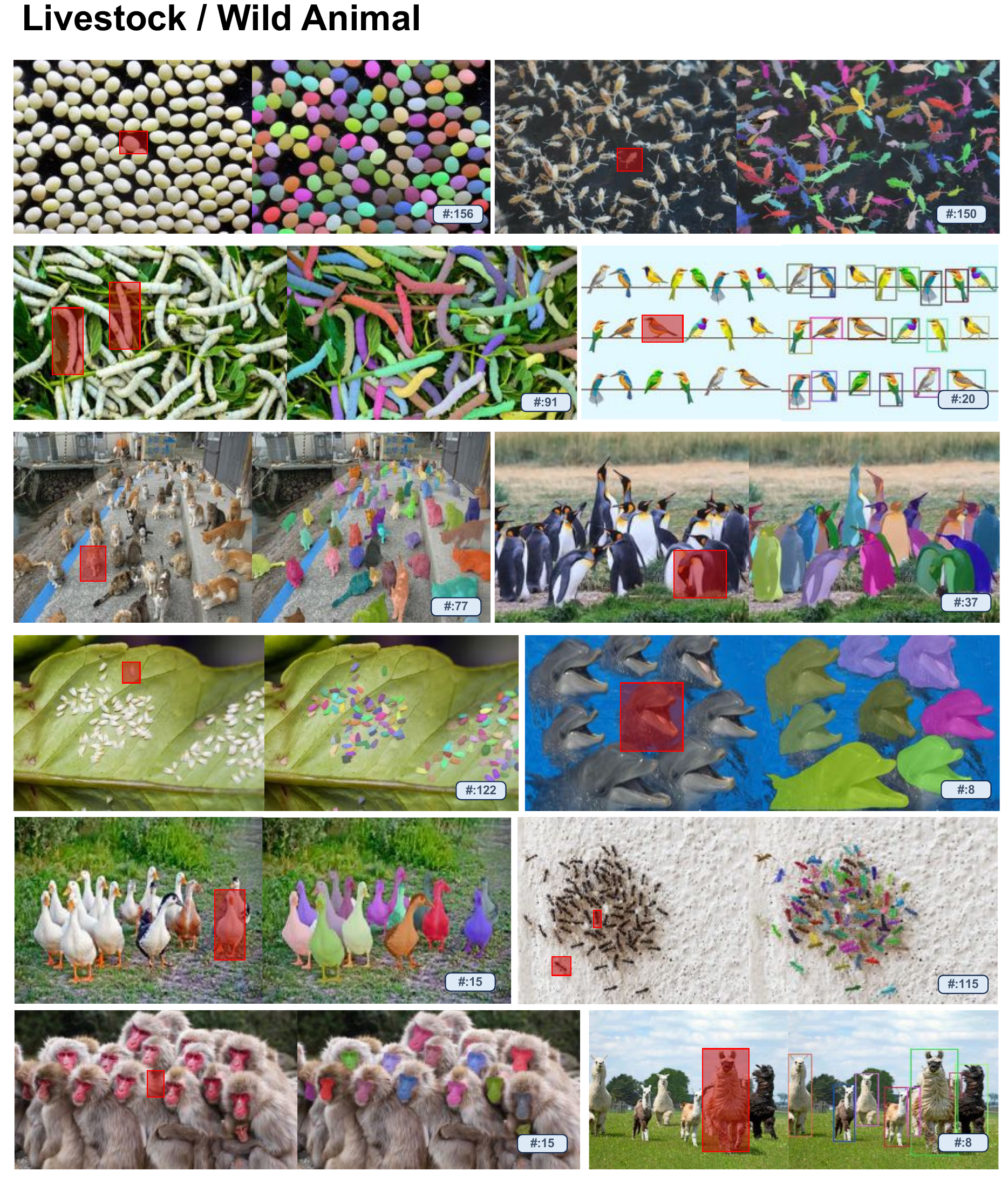}\vspace{-1mm}
\caption{T-Rex applied to livestock and wild animal. To more intuitively visualize overlapping and dense objects, we prompt SAM with the boxes predicted by T-Rex to obtain segmentation masks.}
\label{fig:animal2}
\vspace{-1mm}
\end{figure*}

\begin{figure*}[t]\centering
\includegraphics[width=0.99\linewidth]{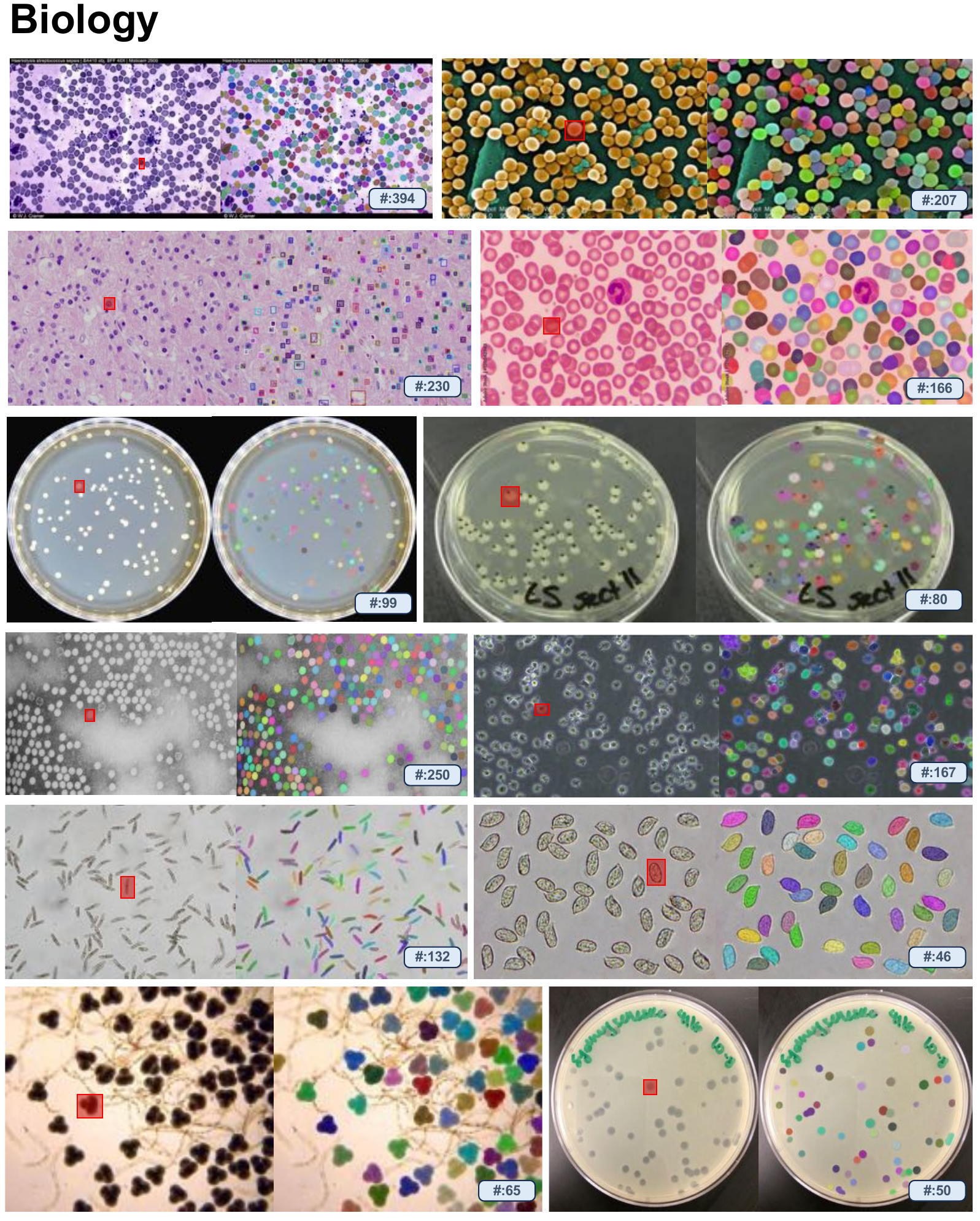}\vspace{-1mm}
\caption{T-Rex applied to biology. To more intuitively visualize overlapping and dense objects, we prompt SAM with the boxes predicted by T-Rex to obtain segmentation masks.}
\label{fig:biology}
\vspace{-1mm}
\end{figure*}

\begin{figure*}[t]\centering
\includegraphics[width=0.90\linewidth]{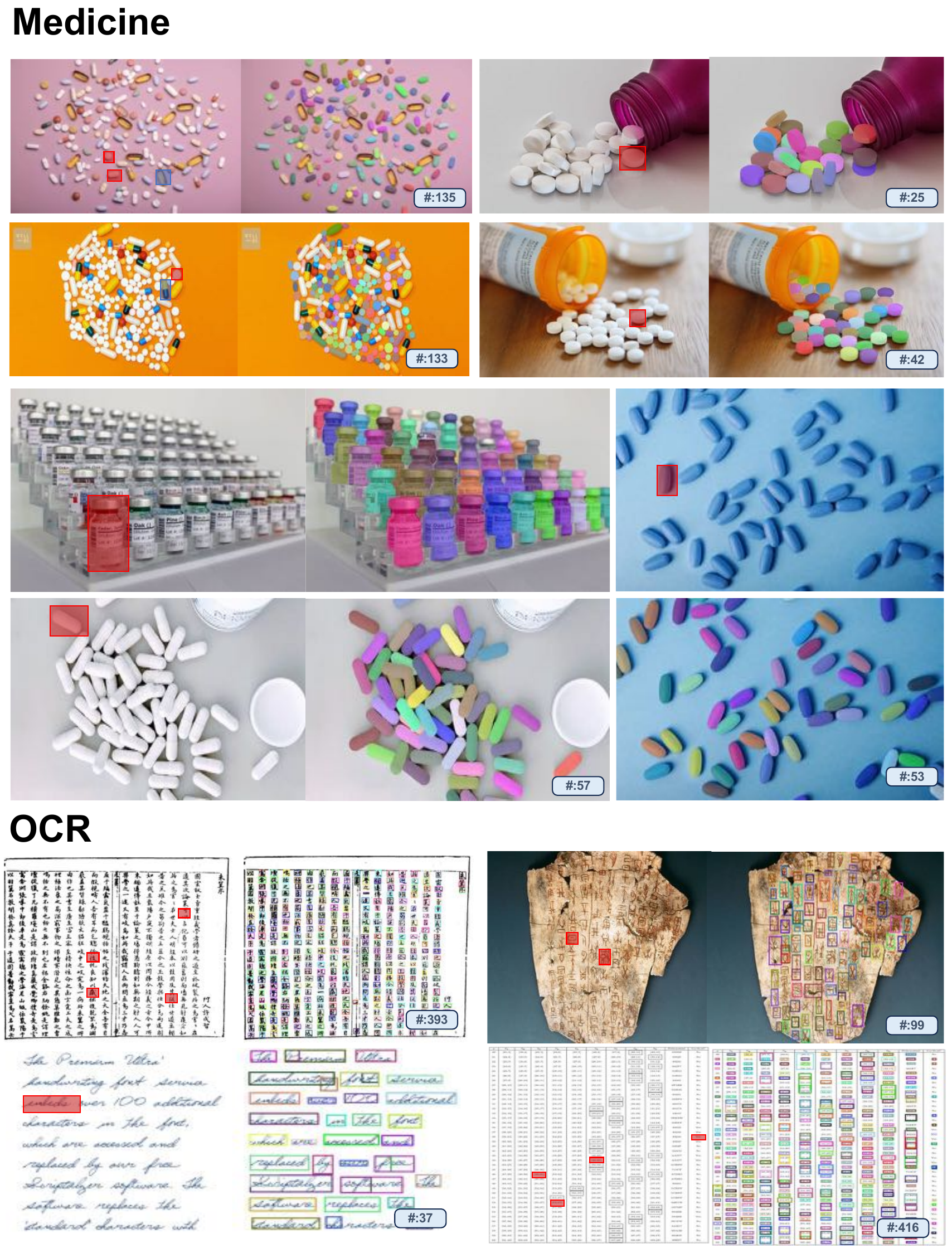}\vspace{-1mm}
\caption{T-Rex applied to medicine and OCR. To more intuitively visualize overlapping and dense objects, we prompt SAM with the boxes predicted by T-Rex to obtain segmentation masks.}
\label{fig:medicine}
\vspace{-1mm}
\end{figure*}

\begin{figure*}[t]\centering
\includegraphics[width=0.99\linewidth]{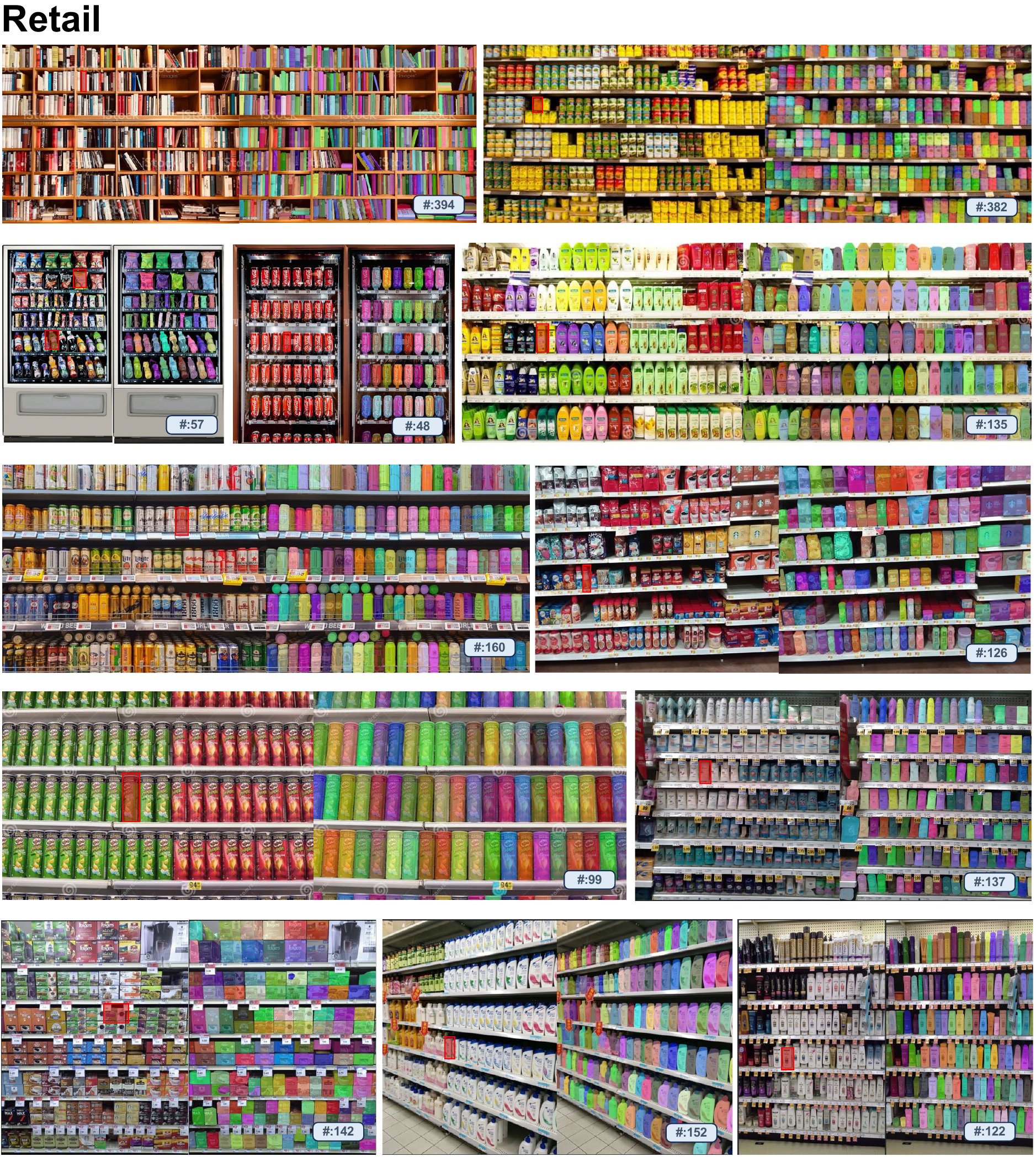}\vspace{-1mm}
\caption{T-Rex applied to retail. To more intuitively visualize overlapping and dense objects, we prompt SAM with the boxes predicted by T-Rex to obtain segmentation masks.}
\label{fig:retail}
\vspace{-1mm}
\end{figure*}

\begin{figure*}[t]\centering
\includegraphics[width=0.99\linewidth]{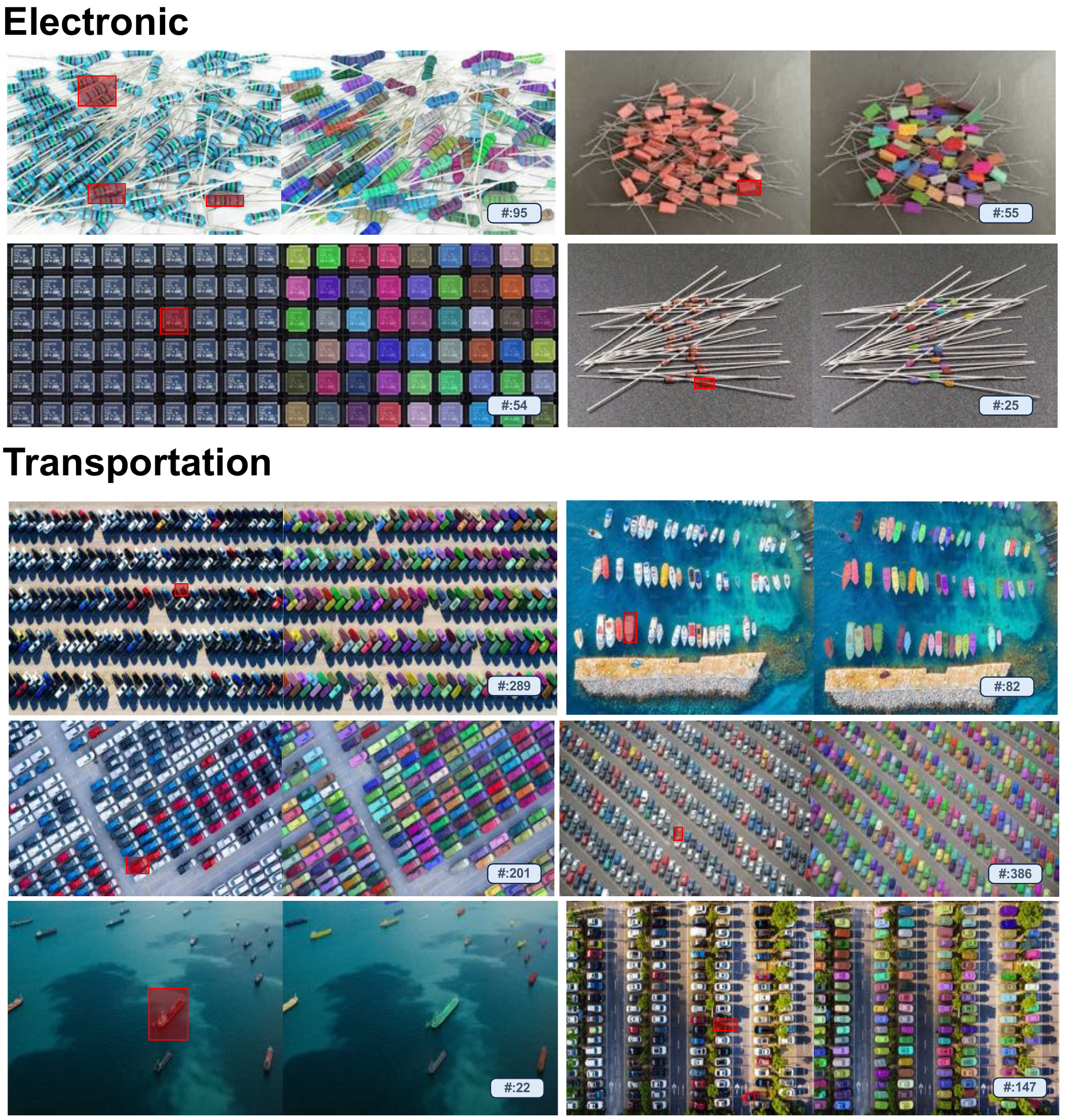}\vspace{-1mm}
\caption{T-Rex applied to electronic and transportation. To more intuitively visualize overlapping and dense objects, we prompt SAM with the boxes predicted by T-Rex to obtain segmentation masks.}
\label{fig:electronic}
\vspace{-1mm}
\end{figure*}

\begin{figure*}[t]\centering
\includegraphics[width=0.99\linewidth]{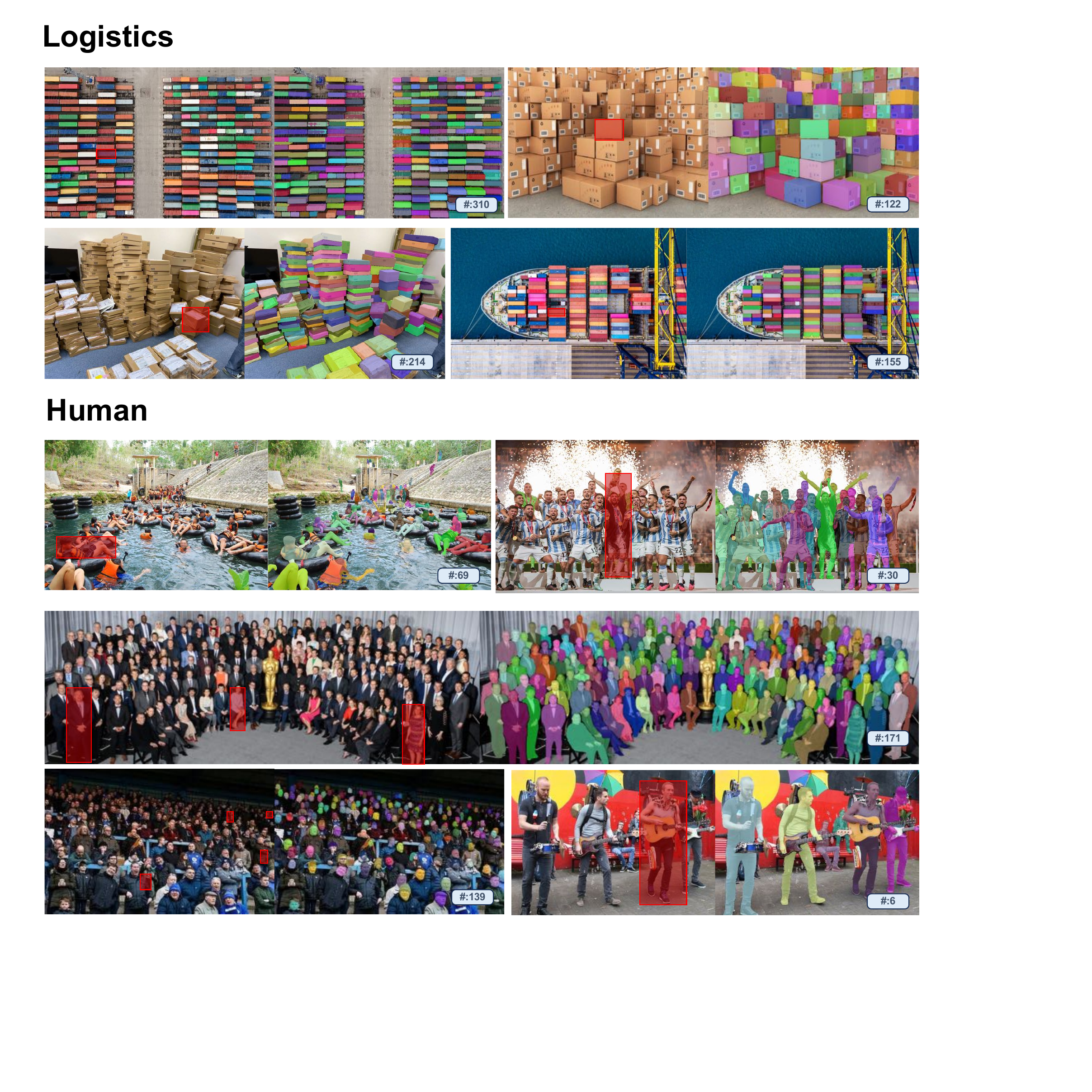}\vspace{-1mm}
\caption{T-Rex applied to logistics and human. To more intuitively visualize overlapping and dense objects, we prompt SAM with the boxes predicted by T-Rex to obtain segmentation masks.}
\label{fig:logistics}
\vspace{-1mm}
\end{figure*}

\begin{figure*}[t]\centering
\includegraphics[width=0.99\linewidth]{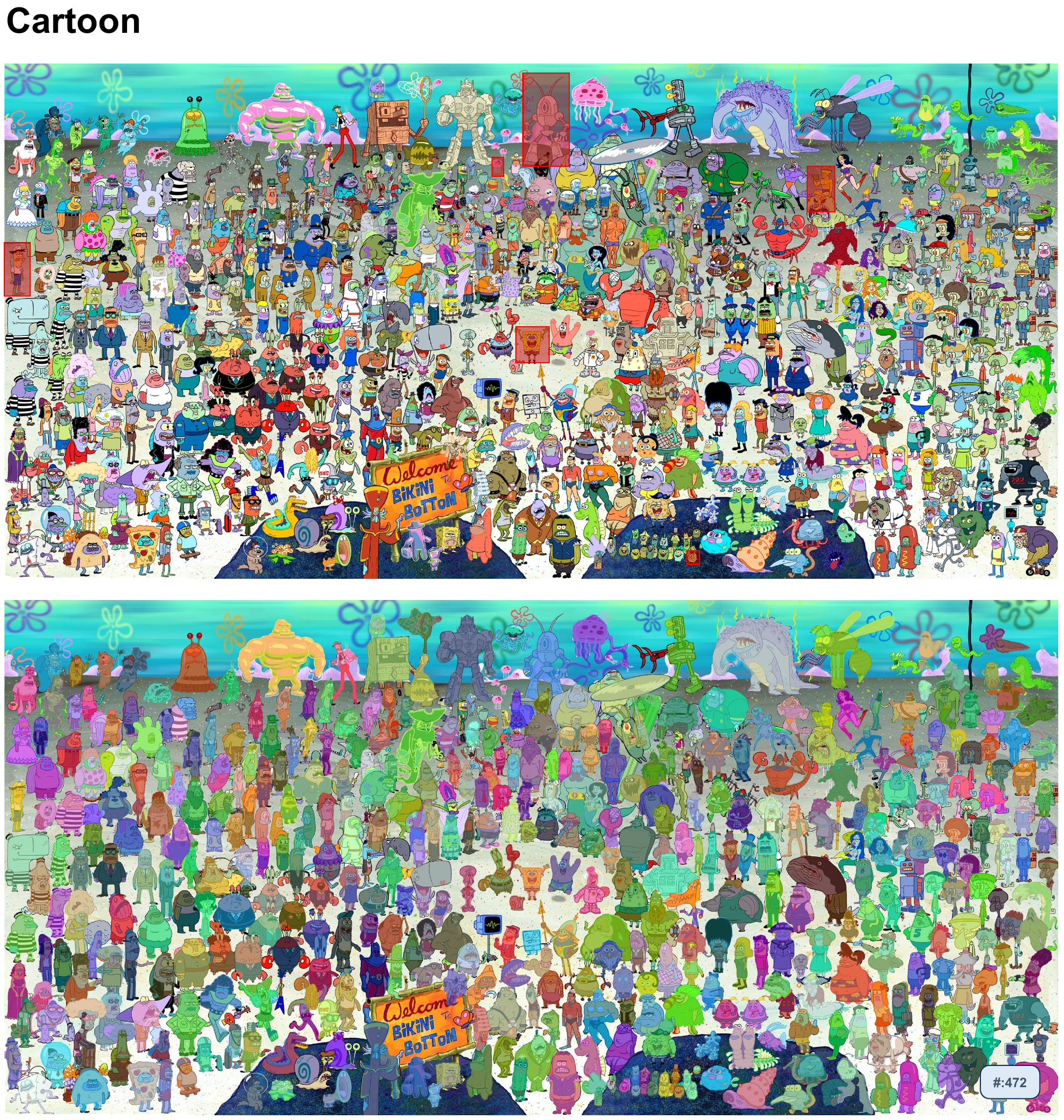}\vspace{-1mm}
\caption{T-Rex applied to cartoon. To more intuitively visualize overlapping and dense objects, we prompt SAM with the boxes predicted by T-Rex to obtain segmentation masks.}
\label{fig:cartoon}
\vspace{-1mm}
\end{figure*}

\begin{itemize}
    \item Agriculture: Fig. \ref{fig:agriculture}.
    \item Industry: Fig. \ref{fig:industry1}, Fig. \ref{fig:industry2},  Fig. \ref{fig:industry3}.
    \item Livestock and Wild Animal: Fig. \ref{fig:animal}, Fig. \ref{fig:animal2}.
    \item Biology: Fig. \ref{fig:biology}.
    \item Medicine: Fig. \ref{fig:medicine}.
    \item  OCR: Fig. \ref{fig:medicine}.
    \item Retail: Fig. \ref{fig:retail}.
    \item Electronic: Fig. \ref{fig:electronic}.
    \item Transportation: Fig. \ref{fig:electronic}.
    \item Logistics: Fig. \ref{fig:logistics}.
    \item Human: Fig. \ref{fig:logistics}.
    \item Cartoon: Fig. \ref{fig:cartoon}.
\end{itemize}

{
    \small
    \bibliographystyle{ieeenat_fullname}
    \bibliography{main}
}

